\documentclass{article}

\usepackage{microtype}
\usepackage{graphicx}
\usepackage{subfigure}
\usepackage{amsmath}
\usepackage{amsfonts}
\usepackage[T1]{fontenc}
\usepackage{multirow}
\usepackage{footnote}
\usepackage{commath}
\usepackage{algorithm}
\usepackage{siunitx}
\DeclareMathOperator*{\argmin}{argmin}

\usepackage{booktabs} 

\usepackage{hyperref}

\usepackage[accepted]{icml2020}

\icmltitlerunning{Generative Adversarial Simulator}

\begin{document}

\twocolumn[
\icmltitle{Generative Adversarial Simulator}




\begin{icmlauthorlist}
\icmlauthor{Jonathan Raiman}{ps}
\icmlaffiliation{ps}{Paris-Saclay}
\end{icmlauthorlist}

\icmlcorrespondingauthor{Jonathan Raiman}{jonathanraiman@gmail.com}

\icmlkeywords{Distillation, Machine Learning, Deep Learning}


\vskip 0.3in
]



\printAffiliationsAndNotice{}  

\newcommand{\methodname}{GAS}
\newcommand{\longmethodname}{Generative Adversarial Simulator}

\begin{abstract}
Knowledge distillation between machine learning models has opened many new avenues for parameter count reduction, performance improvements, or amortizing training time when changing architectures between the teacher and student network.
In the case of reinforcement learning, this technique has also been applied to distill teacher policies to students. Until now, policy distillation required access to a simulator or real world trajectories.

In this paper we introduce a simulator-free approach to knowledge distillation in the context of reinforcement learning. A key challenge is having the student learn the multiplicity of cases that correspond to a given action.
While prior work has shown that data-free knowledge distillation is possible with supervised learning models by generating synthetic examples, these approaches to are vulnerable to only producing a single prototype example for each class. We propose an extension to explicitly handle multiple observations per output class that seeks to find as many exemplars as possible for a given output class by reinitializing our data generator and making use of an adversarial loss.

To the best of our knowledge, this is the first demonstration of simulator-free knowledge distillation between a teacher and a student policy. This new approach improves over the state of the art on data-free learning of student networks on benchmark datasets (MNIST, Fashion-MNIST, CIFAR-10), and we also demonstrate that it specifically tackles issues with multiple input modes. We also identify open problems when distilling agents trained in high dimensional environments such as Pong, Breakout, or Seaquest.
\end{abstract}
\section{Introduction}

The motivation for data-free Knowledge Distillation in the context of Reinforcement Learning (RL) is twofold. Firstly to obtain a compact, portable, student policy from an expert teacher whose network architecture can be changed while retaining high performance on the target environment. This Knowledge Distillation (KD) is of high importance in RL given that model performance is highly variable and often hard to reproduce \cite{henderson2018deep}. Secondly, data-free training: to train the student policy without relying on the original data used to create the teacher. Eliminating the reliance on training data is particularly relevant for RL because they can require thousands of years of experience or 100,000+ CPU cores to train \cite{openai2019dota,andrychowicz2020learning}.

\cite{hinton2015distilling} have shown how Knowledge Distillation (KD) can be used to create student models that obtain higher performance by learning from the teacher's behavior than by training on the original examples. Yet KD requires access to the original data used to train the teacher.

Pioneering work on data-free training \cite{lopes2017data,nayak2019zero,chen2019data} has shown how to generate prototypical examples by back-propagating through the teacher into its inputs. These generated examples can be used to train student models solely through synthetic examples. But as we will show in this paper, such techniques can fail to generate the multiplicity of prototypical examples for a given teacher output class. This creates blind spots resulting in poor student performance. Unfortunately, RL policies must often map multiple observations to the same single desired action, thereby requiring that synthetic data reproduce a wide range of cases for each output response.

We propose \longmethodname\,(\methodname), a new algorithm for data-free learning which explicitly seeks multiple input modes for each output class by combining an adversarial loss and periodically reinitializing the generator. In our experiments we show how these changes enable better coverage of the input modes and increase the accuracy of a student network trained on data that has multiple exemplars per output class.

We demonstrate how we can distill teacher policies into students without a simulator by using our adversarially trained generators. On experiments with RL benchmarks such as Humanoid, MountainCar, CartPole, we show how this technique enables us to distill a teacher. We also perform experiments on the ALE and identify some areas for future work when dealing with high dimensional observation spaces. 
We also validate \methodname\,on three supervised learning data-free KD benchmarks and obtain a new state of the art on MNIST, Fashion-MNIST, and CIFAR-10.

This paper is structured as follows, first we will discuss related work, in Section 2 we will provide a description of the problem, in Section 3 we will describe how the \methodname\,algorithm works, in Section 4 we provide results on our experiments, in Section 5 we discuss these results, and in Section 6 we conclude and provide future work directions.

\section{Related Work}
Our work bridges multiple research areas from knowledge distillation, policy distillation, and imitation learning.

\subsection{Knowledge Distillation}

Knowledge distillation (KD) \cite{hinton2015distilling} is a technique for training a neural network by supervising its outputs using the outputs of another teacher neural network. The ``Dark Knowledge" \cite{hinton2014dark} or additional information contained in the outputs has been used to produce student models that are more compact, achieve higher accuracy than when trained on the original examples, or retain the teacher's accuracy with a noisier feature set \cite{watanabe2017student}.

Our work is most closely related to recent efforts on distilling teachers without access to the original training examples. Initial attempts used meta-data \cite{lopes2017data} collected during training to reconstruct examples and guide KD. In ZSKD the authors show that it is possible to optimize the inputs using a DeepDream-like \cite{mordvintsev2015inceptionism} procedure and generate several thousand synthetic examples that are sufficient to train models on MNIST and Fashion-MNIST with high accuracy relative to students trained with access to the original examples. In DAFL \cite{chen2019data} the authors proprose to instead learn an example generator by using a GAN-like \cite{goodfellow2014generative} objective with the teacher acting as a discriminator, and show that this can be used to achieve good performance on CIFAR-10 and CIFAR-100 as well.

In our work we also train an example generator to perform KD without original examples. Unlike DAFL, we replace the ``one-hot" generator loss by a smoother entropy loss, and use the student during the course of training as a way of informing the generator of areas that require additional supervision. Our training procedure also introduces the use of multiple generators and re-initializations to address vulnerabilities in DAFL when an output has multiple associated proto-examples.

\subsection{Policy Distillation and Imitation Learning}

Our work applies ideas from data-free KD to policy distillation. KD has been used before in a reinforcement learning context as a way of compacting a teacher policy, or combining teachers into a multi-task student \cite{rusu2015policy}. Several techniques have been proposed to blend actions taken by the teacher and student and enable better consolidation and transfer of knowledge \cite{ross2011reduction,ho2016generative,duan2017one}. Until now, policy distillation has relied on access to a simulator.

Imitation learning is another technique for transferring knowledge to a student policy without a simulator by training offline a student to reproduce expert trajectories. In the work of \cite{abbeel2004apprenticeship} the authors try to find offline a reward function that corresponds to expert trajectories, while more recent work uses large amounts of human Go and StarCraft 2 games to bootstrap policies  \cite{silver2016mastering,vinyals2019alphastar,vinyals2019grandmaster}.

Our proposed approach most resembles work on Policy Distillation \cite{rusu2015policy} but removes the need for a simulator or expert trajectories by replacing rollout data by synthetic examples from a generator.

\section{Problem Statement}
\subsection{Overview and Notation}

We are given a {\it teacher} function $f_{\mathrm{t}}(x;\theta_{\mathrm{t}}) \to \hat{y}$, parametrised by $\theta_t$ (for brevity we will omit this term), trained with labeled inputs from an unknown but predetermined teacher training distribution $(x, y) \sim (X, Y), X \subset \mathbb R^n, Y \subset R^m$. We know the teacher minimizes some distance $D$ between $\hat{y}$ and the label $y$: $\theta_{\mathrm{t}}^* = \argmin_{\theta_t}\left\{\mathbb E\left[D(f_{\mathrm{t}}(x;\theta_{\mathrm{t}}), y)\right]\right\}$, we want to obtain a new {\it student} function $f_{\mathrm{s}}(x;\theta_{\mathrm{s}})$, parametrised by $\theta_s$, which minimizes the expected distance $D$ between the outputs of $f_t$ and $f_s$:
\begin{align}
\begin{split}
\mathcal L_{\mathrm{distill}}(\theta_s, X) &= \mathbb E_{x \sim X} \left [D(f_{\mathrm{s}}(x;\theta_{\mathrm{s}}), f_{\mathrm{t}}(x))\right]\\
\theta_s^* &= \argmin_{\theta_s}\left\{\mathcal L_{\mathrm{distill}}(\theta_s, X)\right\}.
\end{split}\label{eq:distill}
\end{align}
Finding $\theta_s^*$ is difficult because the teacher training distribution $(X,Y)$ is unknown. A brute force approach could involve sampling $N$ random points from $\mathbb R^n$ and iteratively minimizing $\mathcal L_{\mathrm{distill}}(\theta_s, \mathbb R^n)$ with respect to $\theta_s$, and as $N\to\infty$, we expect to $\mathcal L_{\mathrm{distill}}(\theta_s, X)$ to be minimized as well because $X \subset \mathbb R^n$. However the brute force approach faces several difficulties identified in prior work \cite{lopes2017data,nayak2019zero,chen2019data}: 1) for large $n$, it may require an intractable amount of points $N$ to minimize $\mathcal L_{\mathrm{distill}}(\theta_s, X)$, 2) the student function $f_s$ may not have the same expressive capacity as $f_t$, and thus sampling points at random might over-subscribe the student, while limiting the capacity over the points of interest.
\subsection{Simplifying Assumptions}
\label{section:generatorloss}

To make this problem tractable, some simplifying assumptions have been made in prior work about the output space $Y$ and the distance $D$: if $Y$ is a simplex, and the teacher was trained to minimize the Kullback-Leibler divergence with respect to vertices of this simplex (e.g. categorical class labels), then we can improve over the brute force approach by only sampling points $x_{\mathrm{fake}}\sim X_{\mathrm{fake}}$ that minimize the entropy $\mathbb E_{x_{\mathrm{fake}}\sim X_{\mathrm{fake}}}\left[H(f_{\mathrm{t}}(x_{\mathrm{fake}}))\right]$. We also assume that $Y$ contains all the vertices of the simplex (e.g. the output spans all output classes), therefore we want to ensure that we maximize the entropy of the expected output: $H\left(\mathbb E\left[f_{\mathrm{t}}(x_{\mathrm{fake}}\right]\right)$.

Because $f_t$ is not necessarily invertible, we now have a second optimization problem, with hyperparameters $\alpha$, $\beta$, find a distribution $X_{\mathrm{fake}}$ that minimizes:
\begin{align}
\begin{split}
\mathcal L_{H} =& \alpha \cdot \mathbb E_{x_{\mathrm{fake}} \sim X_{\mathrm{fake}}}\left[H(f_t(x_{\mathrm{fake}}))\right]\\
&- \beta \cdot H(\mathbb E_{x_{\mathrm{fake}}\sim X_{\mathrm{fake}}}\left[f_t(x_{\mathrm{fake}})\right]).
\end{split}\label{eq:generatorlossentropy}
\end{align}
In the DAFL approach \cite{chen2019data} the authors also add an ``activation loss" weighed by $\lambda_{\mathrm{activation}}$, where they assume that $x_{\mathrm{fake}}\sim X_{\mathrm{fake}}$ should maximize the mean absolute value of the last hidden layer $h_t(x)$ of the (neural network) teacher. Combining these assumptions gives us the following non-adaptive generator loss:
\begin{align}
\mathcal L_{G} =& \mathcal L_{H} - \lambda_{\mathrm{activation}} \cdot \mathbb E_{x_{\mathrm{fake}} \sim X_{\mathrm{fake}}}\left[\abs{h_t(x_{\mathrm{fake}})}\right].\label{eq:generatorloss}
\end{align}

\subsection{Relation to Reinforcement Learning}

Unlike prior work, we are also interested in distilling teachers that were trained through reinforcement learning by interacting with an environment and using a policy gradient algorithm to learn the parameters to maximize the expected sum of future discounted rewards. Because teacher training is driven by environment interaction, there is no associated fixed teacher training distribution such as MNIST or CIFAR-10. In other regards, the teacher functions learnt by policy gradient (commonly named policies $\pi_t$) are functionally identical to the teacher functions $f_t$ described above: $\pi_t$ maps a state $x$ to a distribution $y$, from which an action $a$ is chosen.

Some of our earlier assumptions still hold in the reinforcement learning setting: the policy $\pi_t$ will assign higher probability to actions $a$ that maximize expected discounted future reward. If this policy has no entropy regularization, then the entropy of its action distribution will drop\footnote{We assume the environment is stochastic but fixed, unlike for instance a two-player competitive environment where the opponent might adapt to the current player.}. We can assume the points of interest $x_{\mathrm{fake}}$ minimize $\mathbb E_{x_{\mathrm{fake}}}\left[H(f_t(x_{\mathrm{fake}}))\right]$ and if all actions can be sampled then we should also maximize $\mathbb H_{x_{\mathrm{fake}}}(E\left[f_t(x_{\mathrm{fake}})\right])$, therefore our criteria for $X_{\mathrm{fake}}$ from \eqref{eq:generatorloss} remains applicable.

Our assumption that $Y$ is a simplex may not always be true: policies commonly have continuous action spaces, thus $Y$ is either a probability distribution function or a probability mass function. Furthermore, we previously assumed the teacher's training distribution $(X, Y)$ was predetermined, however in a stateful environment the teacher training distribution is path dependent: $X_t$, the state distribution after $t$ actions are taken, is a function of the previous state $x_{t-1}$ and the previous actions $\{a_0,\dots,a_{t-1}\}$. For this reason certain states of the world are only reachable if the policy takes a specific sequence of actions (such as finding a key to open a door). This dependency means that improvements in the overall distillation objective from \eqref{eq:distill} may not always translate into gains in expected rewards in the student policy since certain sequences of actions matter more than others.

\subsection{Space Invertibility}
\label{section:invertibility}
\begin{figure*}[ht]
\centering
\includegraphics[height=1.7in]{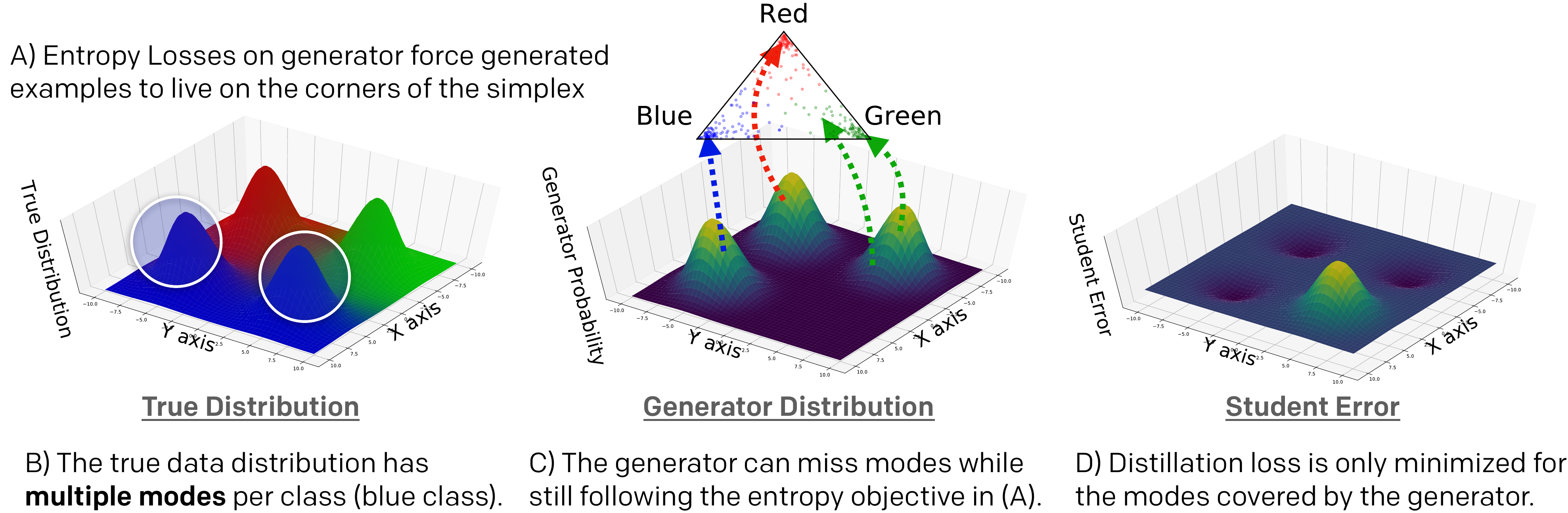}
    \caption{Toy example with 2D data from 3 classes. 
    A generator can minimize the entropy losses despite missing input modes. An adaptive loss biases the generator towards the region where the student disagrees with the teacher, ``seeking" out missing modes.}
    \label{fig:loss_illustration}
\end{figure*}
The output space of the teacher is not always invertible leading to issues with generators only creating a single proto-example per output class. An explanation for this problem is that the teacher function provide a mapping: $f_t: \mathbb R^n \mapsto \mathbb R^m$, where there might exist inputs $x_1, x_2$, $x_1 \neq x_2$ with $f_t(x_1) = f_t(x_2)$, especially when the output space is a simplex, and the teacher was trained with examples that are simplex vertices (categorical labels). This surjectivity lets distribution $X_{\mathrm{fake}}$ that contain only a subset of the input modes but maximize the entropy of the teacher output $H(\mathbb E_{x_{\mathrm{fake}}\sim X_{\mathrm{fake}}}\left[f_t(x_{\mathrm{fake}})\right])$ to minimize the objective $\mathcal L_H$ \eqref{eq:generatorlossentropy}.

We illustrate this problem in Figure~\ref{fig:loss_illustration} where a teacher is trained to classify points originating from 4 quadrants of $\mathbb R^2$ into 3 classes. The distribution shown in part C) ignores one of the input modes but is still a minima of $\mathcal L_H$ \eqref{eq:generatorlossentropy}. As a result, the student might produce outputs that disagree with the teacher when sampling from the true distribution (part D) because it never observed data from the missing mode.

\section{Approach}

\subsection{Goal}

We perform KD as introduced in \cite{hinton2014dark} in a setting where we do not have access to the original examples used to train the teacher \cite{lopes2017data,nayak2019zero,chen2019data}. Our goal is to learn the parameters $\theta_s$ of a student function $f_s(x; \theta_s) \to \hat{y}$ to minimize the distillation loss defined in \eqref{eq:distill} given our teacher function $f_t(x)$. We propose to train a \longmethodname\, (GAS), a model that generates synthetic examples that are maximally informative when distilling the teacher's knowledge into the student. Unlike prior work on data-free KD that is vulnerable to having only a single proto-example per teacher output, our approach uses an adversarial loss, re-initializations, and multiple generators to seek as many examples as possible for each output class.

\subsection{Training objectives}

We supervize our student $f_s$ by sampling points from a surrogate distribution $x_{\mathrm{fake}}\sim X_{\mathrm{fake}}$ and let the teacher label examples $\left(x_{\mathrm{fake}}, f_t(x_{\mathrm{fake}})\right)$. Since $\mathcal L_{\mathrm{distill}}$ is differentiable with respect to the student parameters $\theta_s$, we can perform stochastic gradient descent to learn $\theta_s$.

To obtain student training data we must construct a source of synthetic examples $X_{\mathrm{fake}}$. Similar to \cite{chen2019data}, our $X_{\mathrm{fake}}$ is a learnt generator that minimizes the objective defined in \eqref{eq:generatorloss}. In Section \ref{section:generatorloss} we motivated the use of entropy to obtain a generator that biases the sample generation towards a region of high teacher confidence, as well as ensuring that the generated training data has outputs that span the entire output space (e.g. if $f_t$ outputs a probability mass function, then cover all the vertices of the simplex $Y$).

Generators $X_{\mathrm{fake}}$ that minimize $\mathcal L_{G}$ are vulnerable to selecting distributions that only have a single prototype for each output class as we explain in Section \ref{section:invertibility}. We propose to extend the generator's objective to explicitly handle multiple input modes by adding a term that biases the generation towards regions of the space where the teacher and the student disagree. As illustrated in Figure~\ref{fig:loss_illustration}, a non-adaptive generator distribution produces samples that correspond to vertices of output space simplex $Y$. We make the generator objective also maximize student error (D in Figure~\ref{fig:loss_illustration}), thereby biasing the generation towards areas of the input space that both minimize the entropy of the output, $H(f_t(x_{\mathrm{fake}}))$, while also maximize the distance between the student and teacher output. Formally, the adaptive generator loss combines \eqref{eq:distill} and \eqref{eq:generatorloss} with a hyperparameter $\gamma$:
\begin{align}
\mathcal L_{\mathrm{adapt}}(\theta_s) &= \mathcal L_G - \gamma \mathcal L_{\mathrm{distill}}\left(\theta_s, X_{\mathrm{fake}}\right). \label{eq:adaptiveloss}
\end{align}

\subsubsection*{Probability Distribution Functions}

Computing $\mathcal L_H$ presents a practical challenge in a reinforcement learning setting with a continuous action space policy because we do not always have a closed form expression for $H(\mathbb E_{x_{\mathrm{fake}}\sim X_{\mathrm{fake}}}\left[f_t(x_{\mathrm{fake}})\right])$.
In this work we restrict ourselves to diagonal Gaussian output spaces, making $H\left(\mathbb E_{x_{\mathrm{fake}}\sim X_{\mathrm{fake}}}\left[f_t(x_{\mathrm{fake}})\right]\right)$ be the entropy of a Gaussian mixture,
that we approximate using the closed-form lower bound given in \cite{huber2008entropy}. We reproduce it below, with $b$ the batch size:
\begin{align}
\begin{split}
H_{\mathrm{lower}}(x) &= -\sum_{i=1}^b w_i \cdot \log\left(\sum_{j=1}^b w_j \cdot z_{i,j}\right),\\
\text{with}\,z_{i,j} &= \mathcal N(\mu_i ; \mu_j, C_i + C_j).
\end{split}
\end{align}

\subsection{Neural Network Architecture}

\methodname\,places little requirement on the kinds of neural network architectures used by the teacher and student models. Students and teachers $f_t(x)$ must have outputs that are differentiable with respect to the inputs, and the input and output dimensions for both models must be identical.

Generators in our work change architecture based on the teacher: in image domains we use a DCGAN or another architecture that has transposed-convolutions but does not use Batch Norm \cite{ioffe2015batch}, while in domains without spatial structure we use a deep neural network with Relu nonlinearities but other architecture might work just as well. In both domains the generator receives as input a noise vector, following the approach taken in DAFL \cite{chen2019data}. Moreover, to increase the diversity of the generated outputs we can train multiple generators in parallel that share a single student and teacher. Our noise vector batch is divided among the multiple generators and their outputs are concatenated back.

\subsection{Algorithm}

\methodname\,involves two losses that must be optimized individually: distillation $\mathcal L_{\mathrm{distill}}$ \eqref{eq:distill} and an adaptive loss $\mathcal L_{\mathrm{adapt}}$ \eqref{eq:adaptiveloss}. At the beginning of training, the adaptive loss is not informative because the student is untrained, however when the distance between the student and teacher's output shrinks, it can then be used to improve the generator by identifying missing inputs. We alternate optimization between the student's distillation loss and the generator's adaptive loss.

Because the generator's solution space is very large, it is possible that even with an adaptive loss it does not successfully explore the full input space. We propose to periodically randomly reinitialize weights of the generator to force exploration.

We observe a similar instability phenomenon to GAN training \cite{goodfellow2014generative,radford2015unsupervised,Mescheder2018ICML} when using an adaptive loss in our generator. For this reason, unlike DAFL that interleaves generator and student training, we alternate the losses and perform a different number of gradient steps on the generator and the student. The exact training procedure for \methodname~is shown in Algorithm~\ref{alg:algo}. 

\begin{algorithm}[tb]
   \caption{\longmethodname}
   \label{alg:algo}
\begin{algorithmic}
   \STATE {\bfseries Input:} teacher model $f_t(x), f_t: X \mapsto Y, X \subset \mathbb R^n$.
   \STATE Construct generator $g(z, \theta_{g,0})$, Set $i=0$.
   \STATE Construct student $f_s(x, \theta_{s,0})$, Set $j = 0$.
   \FOR{$e=0$ {\bfseries to} $E_s$}
   \FOR{$step=0$ {\bfseries to} $N_s$}
   \STATE Sample fake inputs: $\{g(z_1),\dots,g(z_b)\}$.
   \STATE Get activations: $\{f_t(g(z_1)),f_s(g(z_1)),\dots\}$.
   \STATE Update student: $\theta_{s, j+1} = \mathrm{Adam}(\nabla_{\theta_{s,j}} \mathcal L_{\mathrm{distill}})$.
   \STATE Set $j = j + 1$.
   \ENDFOR
   \IF{$e \mod R_g $ is $0$}
   \STATE Reinitialize generator weights $\theta_{g, i}$.
   \ENDIF
   \FOR{$step=0$ {\bfseries to} $N_g$}
   \STATE Sample fake inputs $\{g(z_1),\dots,g(z_b)\}$.
   \STATE Get activations: $\{f_t(g(z_1)),f_s(g(z_1)),\dots\}$.
   \STATE Update generator: $\theta_{g, i+1} = \mathrm{Adam}\left(\nabla_{\theta_{g,i}} \mathcal L_{\mathrm{adapt}}\right)$.
   \STATE Set $i = i + 1$.
   \ENDFOR
   \ENDFOR
\end{algorithmic}
\end{algorithm}
\section{Results}

Our experiments focus on three aspects of the problem: first we will examine how input modes impact student accuracy across different data-free distillation techniques, second we compare the effect of our proposed approach to other algorithms on existing benchmarks for data-free KD, thirdly we attempt to distill without a simulator pretrained policies in a variety of reinforcement learning environments and measure the effect of different distillation algorithms on the score of the distilled student.

\subsection{Input Modes Effect on Distillation}

\begin{figure}[ht]
\centering
\includegraphics[width=\linewidth]{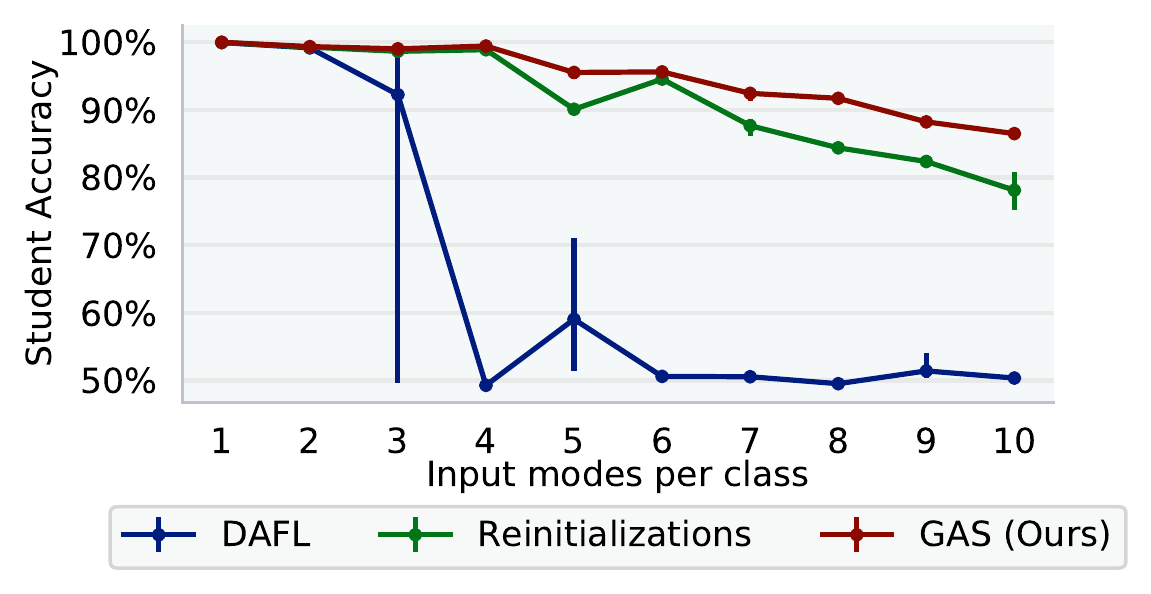}
    \caption{Student accuracy comparison between different data-free KD methods as number of inputs modes per output class is artificially increased. Error bars give 90\% confidence interval.}
    \label{fig:evenodd}
\end{figure}

To isolate the effect of input modes on KD we setup an artificial task where we modify an existing dataset to create experiments where each output class has a set number of input modes. We then train a teacher for each of these datasets and compare the accuracy of students trained with different data-free KD techniques.

Our datasets are constructed by grouping different MNIST digits into a single ``Even" or ``Odd" class. By selecting how many digits we group we can create datasets with 1 to 5 digits per class. We use another MNIST-like dataset (Kuzushiji-MNIST \cite{clanuwat2018deep}) to obtain up to 10 examplars per class.

Our teachers use the LeNet architecture trained using Adam \cite{kingma2014adam} for 300 steps with a learning rate of $7e^{-4}$, and our students use the LeNetHalf architecture with half the filters of the teacher.
The students are trained using DAFL \cite{chen2019data}, a method that uses only re-initilizations and the generative loss $\mathcal L_G$, or using \methodname, which combines re-initilizations, an adaptive loss, and multiple concurrent generators.
Hyperparameters for training the generator and the student are given in appendix \ref{appendix:hyperparam}.

In Figure~\ref{fig:evenodd} we observe the student accuracy as the number of input modes for ``Even" and ``Odd" classes is increased. 
In this plot we notice that past 2 modes per class, the performance of DAFL drops relative to other distillation methods presented. The accuracy of the student decreases with additional input modes. \methodname\, always outperform the other approaches and its accuracy never drops below 85\%.

\subsection{Data-Free Supervised Learning Model Distillation}

\begin{table}[h]
    \centering
    \caption{Classification Accuracy on MNIST dataset}
    \begin{tabular}{l|l}
        Model & Accuracy ($\mu \pm \sigma$)\\
        \bottomrule
        Teacher & 99.35\%\\
        \hline
        \begin{tabular}{@{}l@{}}Student KD \cite{hinton2015distilling}\\{\it 60,000 original examples}\end{tabular} & 98.92\%\\
        \hline
        Meta data \cite{lopes2017data} & 92.47\%\\
        \hline
        DAFL \cite{chen2019data} & 98.20\% \\
        \hline
        ZSKD \cite{nayak2019zero} & 98.77\%\\
        \hline
        \methodname~w/. $\lambda_{\mathrm{activation}}=0.1$ & 
         $98.67\%\pm 0.21$\\
        \hline
        \methodname~(Ours) & {\bf 98.79\%} $\pm 0.075$
    \end{tabular}
    \label{table:mnist}
\end{table}

\begin{table}[h]
    \centering
    \caption{Classification Accuracy on Fashion-MNIST}
    \begin{tabular}{l|ll}
        Model & \begin{tabular}{@{}l@{}}Accuracy \\($\small \mu \pm \sigma$)\end{tabular} & $\frac{\text{Student}}{\text{Teacher}}$\\
        \bottomrule
        Teacher & 91.17\%\\
        \hline
        \begin{tabular}{@{}l@{}}Student KD  \cite{hinton2015distilling}\\{\it 60,000 original examples}\end{tabular} & 89.66\% & 98.70\% \\
        \hline
        \begin{tabular}{@{}l@{}}\cite{kimura2019few} \\{\it 200 original examples} \end{tabular} & 72.50\%\\
        \hline
        ZSKD \cite{nayak2019zero} & 79.62\% & 87.64\%\\
        \hline
        \methodname~(Ours) & \begin{tabular}{@{}l@{}}{\bf81.33\%} \\$\small \pm 0.42$\end{tabular} & \begin{tabular}{@{}l@{}}{\bf89.21\%}\\$\small \pm 0.46$\end{tabular}
    \end{tabular}
    \label{table:fashion}
\end{table}
We investigate whether the use of resetting, adaptive losses, and multiple generators improve the performance when distilling supervised learning models. We follow the same experimental setup as \cite{nayak2019zero,lopes2017data,chen2019data}: we train student models from teachers trained on MNIST \cite{lecun1998mnist}, Fashion-MNIST  \cite{xiao2017fashion}, and CIFAR-10 \cite{krizhevsky2014cifar}.

\subsubsection{Datasets}
We use three different data-free KD benchmark datasets: MNIST, Fashion-MNIST, and CIFAR-10.
The MNIST \cite{lecun1998mnist} dataset contains 60,000 training examples and 10,000 test examples showing handwritten digits from 0 to 9. Fashion-MNIST \cite{xiao2017fashion} dataset was proposed as an alternative to MNIST, while retaining the same format of $28\times28$ images with 60,000 training examples, and 10,000 test examples. However, the dataset has been known to be more challenging to learn due to the higher inter-class similarity. CIFAR-10 is an RGB image dataset with slightly larger images ($32\times 32 \times 3$). It is composed of 50,000 training examples and 10,000 test examples.

\subsubsection{MNIST and Fashion MNIST}
We study data-free distillation performance by first training a teacher on this MNIST and Fashion-MNIST. Our teachers are trained using an Adam \cite{kingma2014adam} optimizer with learning rate 0.0007, and a Dropout \cite{srivastava2014dropout} probability of 0.2. We save the best performing model with a test accuracy of 99.35\% on MNIST and 91.17\% on Fashion-MNIST. We freeze the teachers and use them to compute generator and distillation losses. To be comparable to prior work we adopt a a LeNet-5 \cite{lecun2015lenet} architecture for the teacher, and the student has a LeNet-5-Half architecture (number of filters is halved). Our generator uses the DCGAN architecture from \cite{radford2015unsupervised}, which was also used in DAFL \cite{chen2019data}.

Generators are trained with the adaptive loss from \eqref{eq:adaptiveloss}. At each example generation stage we sample a batch of 512 noise vectors $z \in \mathbb R^{100}$, $z_i \sim \mathcal U(-1, 1)$. These vectors are sliced into 64 sub-batches fed into each of the 8 concurrent generators. We reinitialize the generator once every 4,000 student updates on MNIST, and once every 120,000 student updates on Fashion-MNIST. After each 100 updates to the student, the generators are optimized for 20 steps. The generator and student are trained using the Adam \cite{kingma2014adam} optimizer, with a generator learning rate of 0.01 and a student learning rate of 0.001. We repeat our experiment with 3 seeds and report the mean and standard deviation.
\paragraph{MNIST}
In Table~\ref{table:mnist} we present the performance of several models: the teacher model; Student KD, a model trained using KD with access to the original training data following \cite{hinton2015distilling}; Meta data, an approach that uses meta data from the teacher during training; \cite{lopes2017data}; and two data-free distillation approaches, ZSKD \cite{nayak2019zero}, which uses data-impressions as inputs, and DAFL \cite{chen2019data}, an approach that also trains a generator, but uses a different loss and training algorithm from ours; and our approach, \methodname~with and without an activation loss. Unlike \cite{chen2019data}, we find that activation losses do not help the generator. We note that our approach outperforms DAFL, and slightly outperforms ZSKD. Despite not using any training data, \methodname\, obtains similar performance to Student KD, which uses the original examples.
\paragraph{Fashion-MNIST}
In Table~\ref{table:fashion} we compare the accuracy of the student obtained through our approach, \methodname, to the accuracy from others models. Student KD \cite{hinton2015distilling} is trained on the original examples with labels given by the teacher. In \cite{kimura2019few} the authors use a generative process and a limited number of real examples to train a student. In ZSKD \cite{nayak2019zero} the student is trained using data-impressions without any original examples. To provide a more fair comparison, given that each of these results used a different teacher, we compute the ratio between the student accuracy and the associated teacher accuracy ($\frac{\text{Student}}{\text{Teacher}}$). \methodname\,outperforms techniques with access to few training examples and other data-free KD approaches such as ZSKD by a large margin.

\subsubsection{CIFAR-10}

\begin{table}[h]
    \centering
    \caption{Classification Accuracy on CIFAR-10 dataset}
    \begin{tabular}{l|ll}
        Model & Accuracy & $\frac{\text{Student}}{\text{Teacher}}$\\
        \bottomrule
        Teacher & 95.36\%\\
        \hline
        \begin{tabular}{@{}l@{}}Student KD  \cite{hinton2015distilling}\\{\it 50,000 original examples}\end{tabular} & 94.34\% & 98.70\% \\
        \hline
        DAFL \cite{nayak2019zero} & 92.22\% & 96.48\%\\
        \hline
        \methodname~(Ours) & {\bf 92.35\%} & {\bf 96.84\%}\\
    \end{tabular}
    \label{table:cifar10}
\end{table}

Our setup is identical to the one from \cite{chen2019data}, and we use the implementation released by the authors of \cite{chen2019data}\footnote{\url{https://github.com/huawei-noah/Data-Efficient-Model-Compression}.} to train our teacher, implement our adaptive loss, and train students. In this configuration we use a ResNet-34 \cite{he2016deep} as the teacher, and ResNet-18 \cite{he2016deep} as the student.
The student is trained using SGD with momentum 0.9, and a learning rate of 0.1 that is dropped by a factor of 10 every 96,000 gradient steps, for a total of 240,000 steps. The generator is trained using the Adam \cite{kingma2014adam} optimizer with a learning rate of 0.015. We perform 1 generator update for each 10 student updates. We set $\gamma = 0.4$ in the generator loss \eqref{eq:adaptiveloss}.

In Table~\ref{table:cifar10} we compare the accuracy of the student obtained through our approach, \methodname, to the accuracy from others models. Student KD \cite{hinton2015distilling} is trained on the original examples with labels given by the teacher. DAFL \cite{chen2019data} is a model where the student is trained without any original examples by using a learnt generator trained by combining several loses similar to the entropy losses in \eqref{eq:generatorloss}, along with an activation loss. We also include our reproduction of the DAFL result obtained when running our experiments. As with Fashion-MNIST, we also compute the ratio between the student accuracy and the associated teacher accuracy ($\frac{\text{Student}}{\text{Teacher}}$).
We observe that \methodname\,is able to reach an accuracy close to that of a student trained with the original examples, and slightly outperforms DAFL.

\subsection{Simulator-Free Policy Distillation}

Our experiments focus on measuring the score of student policies obtained through simulator-free KD from pretrained teacher policies on a wide range of reinforcement learning environments from classical control benchmarks \cite{duan2016benchmarking} and the Atari Learning Environment (ALE) \cite{bellemare2013arcade}. Our experiments are conducted using OpenAI gym \cite{brockman2016openai} and utilities from OpenAI baselines \cite{dhariwal2017openai}.

\subsubsection{Teacher Policies}
\label{section:rlteacher}

\begin{figure*}
    \centering
    \includegraphics[height=1.1in]{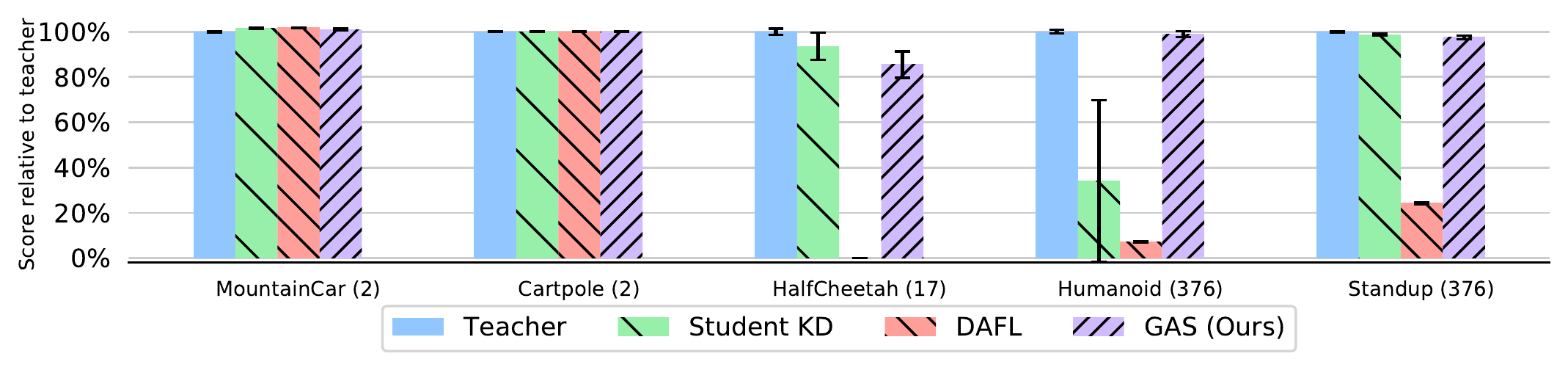}
    \caption{Policy distillation in classical control environments (observation dim). Scores relative to the teachers. Error bars show $\sigma$, $N=5$}
    \label{fig:classicalcontrol}
\end{figure*}

We train policies for the environments listed in Figure~\ref{fig:classicalcontrol} using Proximal Policy Optimization (PPO) \cite{schulman2017proximal} and a Generalized Advantage Estimator (GAE) \cite{schulman2015high} to smooth advantages used to update the policies. This approach has been successfully applied to a wide variety of domains in the past  \cite{bansal2017emergent,baker2019emergent,openai2019dota,andrychowicz2020learning} using both continuous and discrete action spaces. We chose to use a single algorithm for all environments to facilitate reproducibility and simplify comparisons.
Because we only use a single teacher algorithm we do not know the exact impact it can have on the final student's performance, and we leave this investigation as future work. We save the parameters of the best performing teacher policy by measuring its average score over 10 episodes over the course of training. Hyperparameters and additional implementation and technical details are given in Appendix~\ref{appendix:rl}.

\subsubsection{Classical Control}

We study several classical control environments that have continuous (Humanoid, Humanoid-Standup, HalfCheetah, MountainCar) and discrete action spaces (CartPole). The environments have input spaces that vary from $\mathbb R^2$ to $\mathbb R^{376}$, and different dimension output spaces. Using the technique described above we train teachers for all these environments, and then attempt to distill them into student policies.

We compare different KD techniques. Student KD uses real trajectories and minimizes the Kullback-Leibler divergence between the teacher's outputs and those of the student as done in \cite{hinton2015distilling,rusu2015policy}. Using DAFL we use a learnt generator to synthesize observations, and apply the distillation objective given in \cite{chen2019data}. Our approach, \methodname, uses reinitilizations, the same adaptive loss $\mathcal L_{\mathrm{adap}}$ \eqref{eq:adaptiveloss} used with supervised learning models, and a single generator. Hyperparameters for these experiments are given in appendix~\ref{appendix:rldistill}.

In Figure~\ref{fig:classicalcontrol} we present the best reward achieved by these different approaches averaged across 5 seeds. We notice that using real trajectories we are able to learn students that approach the teacher's performance. Using DAFL we are able to learn policies with performance matching the teacher only in the environments with the lowest dimensional input space. Finally, \methodname~is able to learn student policies simulator-free with performance matching the teacher in all environments except HalfCheetah, where we see a slight degradation.

\subsubsection{ALE}

The Atari Learning Environment (ALE) is considered to be a standard benchmark for reinforcement learning algorithms which presents additional challenges regarding exploration and higher dimensional inputs ($160\times210$ RGB frames). We use the Impala \cite{espeholt2018impala} architecture as our teacher policy, and use ``Impala-Half" for our student policy (Impala with half-size filters). We obtain teachers by following the steps in Section~\ref{section:rlteacher}. We preprocess game frames using the same steps used in \cite{espeholt2018impala} so our inputs become a stack of 4 grayscale $84\times84$ frames. The inputs to the teachers and students are normalized using a running mean and standard deviation.

\begin{figure}[ht]
    \centering
    \includegraphics[width=\linewidth]{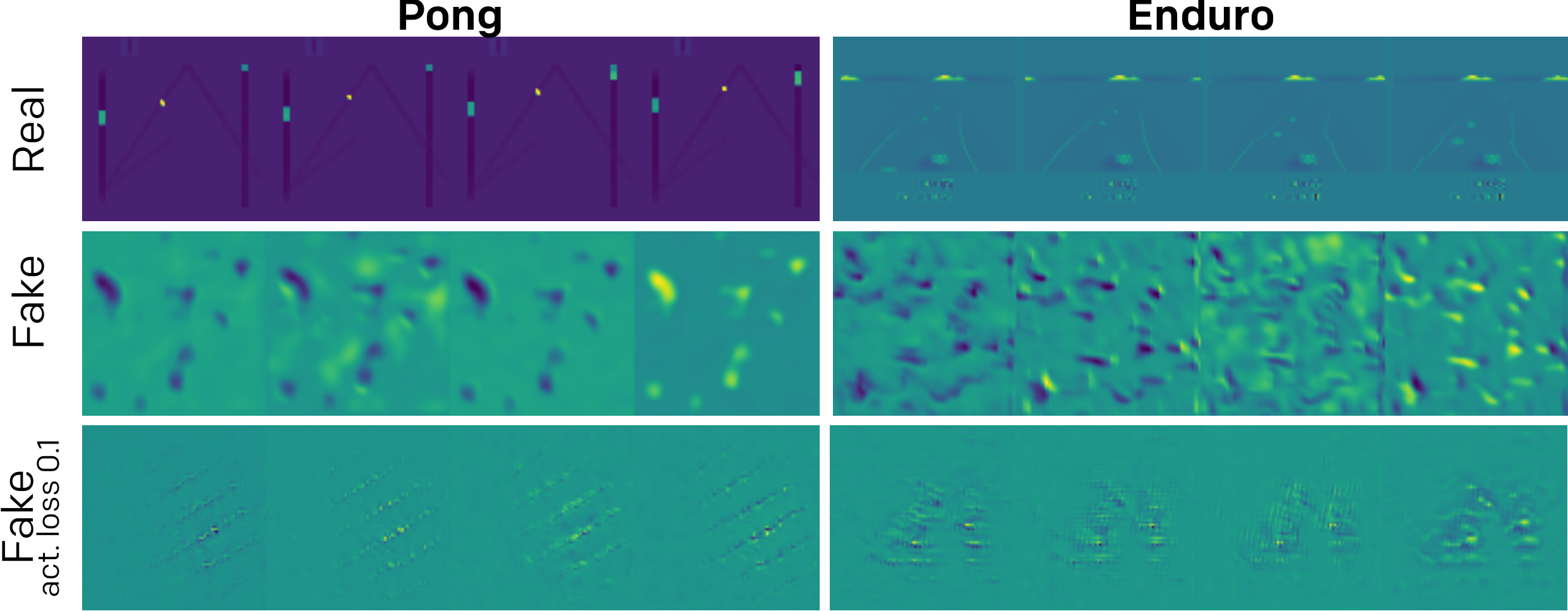}
    \caption{Samples from real inputs and \methodname\, during Enduro and Pong policy distillation. Activation loss leads to recognizable patterns: Enduro road markings and Pong ball trajectories.
    The presence of multiple ball trajectories suggests the generator learns to produce observation superpositions instead of  plausible scenes.
    }
    \label{fig:generations}
\end{figure}

Student KD uses real teacher trajectories collected by multiple concurrent workers to train a policy and minimizes the Kullback-Leibler divergence between the student and teacher outputs following \cite{hinton2015distilling,rusu2015policy}. Our simulator-free approaches, DAFL and \methodname, rely on a learnt generator. Here we use a DCGAN architecture \cite{radford2015unsupervised} to generate  observations. The generated observations skip the input normalization, so the teacher policy expects those inputs to be 0-mean and standard deviation-1.
Both DAFL and \methodname\,fail to learn in these environments. Precise scores in Appendix~\ref{appendix:rl}. With real trajectories Student KD is able to reach performance similar to that of the teacher, confirming the results from \cite{rusu2015policy}. We suspect that the higher dimensional inputs prevented the generators from meaningfully recreating observations that would transfer to the real environment. See Figure~\ref{fig:generations} for generator samples.

We run an additional experiment with a lower dimensional ALE environment to understand if simplifying the generation task can improve \methodname's performance. Generators in the ALE domain produce observations that are $75\times$ larger than those from Humanoid and $9.2\times$ larger than CIFAR-10. We propose to use a 2 stacked frame $48\times48$ Pong environment ($1.5\times$ CIFAR-10) and find that the generations now contain paddles but \methodname's score remains low ($-18.87 \pm 1.17$).
\section{Conclusion}

In this work we considered of data-free KD in the context of reinforcement learning (RL). We identify a vulnerability in existing data-free KD algorithms that can prevent them from providing a multiplicity of examples for each output class and limits their applicability for simulator-free policy distillation.
In an experiment we demonstrate how artificially increasing the number of proto-examples per class lowers the accuracy of the data-free KD technique DAFL.

Our key contribution is \longmethodname\, (\methodname), an algorithm for data-free KD that explicitly seeks to overcome difficulties faced by having multiple examples per output class through the use of adaptive loss, reinitializations, and concurrent generators. We show that this algorithm produces high accuracy students despite increasing input modes.
We validate \methodname\, on standard benchmarks and find that it obtains a new state of the art on existing benchmarks MNIST, Fashion-MNIST, CIFAR-10.

We also introduce the task of simulator-free policy distillation. We train teacher policies on 9 different RL environments to compare policy distillation techniques. On ALE we find that neither DAFL nor \methodname\, successfully trains student policies and can compete with using real trajectories.
On classical control tasks we demonstrate that \methodname\,successfully trains student policies without a simulator. Our approach improves over DAFL in higher dimensional environments, providing some evidence that \methodname\, opens up the possibility to distill policies without simulators in larger and more complex environments. Yet, more work is needed to address higher dimensions and would benefit from from understanding the connection between the teacher-driven observations generations from \methodname\,and those found in World Models \cite{ha2018recurrent} or in model-based RL such as MuZero \cite{schrittwieser2019mastering}.

\bibliography{bibliography}
\bibliographystyle{icml2020}

\appendix
\onecolumn
\section{Hyperparameters}
\label{appendix:hyperparam}

In this section we include the experiment hyperparameters. All experiments are run on a 12-core 3.60GHz Intel i7-6850K CPU with an NVIDIA Titan X Pascal, using Tensorflow 1.15, CUDA~10.1 and CuDNN~7.5.1.

\begin{table}[H]
\caption{Input Mode Effect Hyperparameters}
  \label{table:inputmodehparams}
  \centering
  \begin{tabular}{llll}
  \toprule
  Model & Optimizer & LR & Architecture\\
  \midrule
  Generator & Adam & 0.01 & 
  \small
  DCGAN without Batch-Norm\\
  Student & Adam & 0.001 & LeNetHalf\\
  Teacher & Adam & 0.0007 & LeNet\\
  \end{tabular}
\end{table}

\begin{table}[H]
\caption{Data-Free Distillation Benchmark Hyperparameters}
  \label{table:benchmarkhparams}
  \centering
  \begin{tabular}{llrll}
  \toprule
  Task & Optimizer & Student/Generator LR & Teacher & Student \\
  \midrule
  MNIST & Adam & 0.0001 / 0.01 & LeNet & LeNetHalf\\
  Fashion-MNIST & Adam &  0.0001 / 0.01 & LeNet & LeNetHalf\\
  CIFAR-10 & Adam &  schedule / 0.015 & Resnet34 & Resnet18\\
  \end{tabular}
\end{table}

\subsection{Reinforcement Learning Teachers}
\label{appendix:rl}

We train reinforcement learning agents asynchronously using a single GPU central optimizer machine connected to a series of workers that update their parameters after a set number of transitions. The workers send their observations to a central actor GPU that can batch and process multiple requests. The workers act in parallel with the optimization and regularly send their transition data to a queue on the optimizer machine that normalizes the measure advantage and adds this data to the experience buffer. Data is added to the experience buffer if the rollout was produced within some version lag to avoid straggler workers. Observations for all teachers are normalized using a running mean and standard deviation computed on the optimizer machines. We update these statistics alongside the model parameters, thereby ensuring that the models observe mean 0 and standard deviation 1 data. 

For each environment we train policies until they have stopped improving or we collect more than $10^7$ transitions samples. We save the policy parameters whenever the policy achieves a higher reward across multiple episodes when evaluated.

\begin{table}[H]
\caption{Common Teacher Hyperparameters}
\label{table:commonhyperparams}
  \centering
  \begin{tabular}{llll}
  \toprule
  Parameter & Value\\
  \midrule
  Algorithm & PPO\\
  PPO Clip Ratio & 0.2\\
  Gradient steps per update & 10\\
  Concurrent environments & 512\\
  Entropy penalty & 0.0\\
  Optimizer & Adam\\
  GAE $\lambda$ & 0.95\\
  Concurrent environments & 512\\
  Transitions per rollout & 32\\
  Max version lag & 4\\
  \end{tabular}
\end{table}

\begin{table}[H]
\caption{Environment Specific Teacher Hyperparameters}
  \label{table:envhyperparams}
  \centering
  \begin{tabular}{lrrrrr}
  \toprule
  Environment & Model & Replay Buffer Size & Batch Size & LR & Discount factor ($\gamma$)\\
  \midrule
  Humanoid & MLP([64 64]) & 8192 & 1024 & $3e^{-4}$ & 0.990\\
  Standup & MLP([64 64]) & 8192 & 1024 & $3e^{-4}$ & 0.990\\
  MountainCar & MLP([512]) & 15000 & 5000 & $1e^{-3}$ & 0.995\\
  Cartpole & MLP([512]) & 16384 & 2048 & $1e^{-3}$ & 0.990\\
  ALE & Impala \cite{espeholt2018impala} & 16384 & 2048 & $1e^{-3}$ & 0.990\\
  \end{tabular}
\end{table}

\subsection{Reinforcement Learning Distillation}
\label{appendix:rldistill}

Reinforcement Distillation uses for vector environments (Humanoid, Cartpole, MountainCar) a 100-dimensional uniform [-1, 1] noise vector to drive a single-layer MLP.
For ALE environments we use a DCGAN architecture \cite{radford2015unsupervised}.

\begin{table*}[ht]
    \centering
    \caption{Classical Control scores ($\mu \pm \sigma$, N=5) for student policies.}
    \begin{tabular}{l|
        S[separate-uncertainty,table-figures-uncertainty=1]
        S[separate-uncertainty,table-figures-uncertainty=1]
        S[separate-uncertainty,table-figures-uncertainty=1]
        c
        c}
        Method & {Humanoid} & {Standup ($\times 10^4$)} & {HalfCheetah} & {MountainCar} & {Cartpole}\\
        \bottomrule
        Teacher & 5896 \pm 48.29 & 15.92 \pm 0.042 & 7698 \pm 103.1 &$ 94.30 \pm 0.2547$ & $200.0 \pm 0.000$\\
        Student KD & 2012 \pm 2101. & 15.71 \pm 0.060 & 7199. \pm 459.9 & $95.77 \pm 0.3915$ & $200.0 \pm 0.000$\\
        DAFL & 418.1 \pm 10.75 & 3.854 \pm 0.056 & -0.3870 \pm 0.2094 & $95.87 \pm 0.1889$ & $200.0 \pm 0.000$\\
        \methodname & 5833. \pm 78.38 & 15.52 \pm 0.12 & 6580 \pm 445.3 & $95.13 \pm 0.3705$ & $200.0 \pm 0.000$\\
        \hline
        Input Space & {376} & {376} & {17} & {2} & {4}\\
        Action Space & {$\mathbb R^{17}$} & {$\mathbb R^{17}$} & {$\mathbb R^{6}$} & {$\mathbb R^1$} & {Cat(2)}\\
    \end{tabular}
    \label{table:rl}
\end{table*}

\begin{table*}[ht]
    \centering
    \caption{ALE scores ($\mu \pm \sigma$, N=5) for student policies.}
    \begin{tabular}{l|
        S[separate-uncertainty,table-figures-uncertainty=1]
        S[separate-uncertainty,table-figures-uncertainty=1]
        S[separate-uncertainty,table-figures-uncertainty=1]
        S[separate-uncertainty,table-figures-uncertainty=1]
        }
        Method & {Pong} & {Breakout} & {Seaquest} & {Enduro}\\
        \bottomrule
        Teacher & 20.70 \pm 0.6403 & 355.5 \pm 158.9 & 2624. \pm 552.4 & 643.1 \pm 27.22\\
        Student KD & 20.20 \pm 0.2683 & 213.5 \pm 144.5 & 1015. \pm 577.6 & 119.9 \pm 63.64\\
        DAFL & -20.60 \pm 0.3033 & 0.8400 \pm 0.4800 & 110.0 \pm 15.49 & 4.920 \pm 4.234\\
        \methodname & -20.54 \pm 0.3980 & 0.8800 \pm 0.9683 & 123.6 \pm 21.22 & 24.22 \pm 9.529\\
        \hline
        Action Space & {Cat(6)} & {Cat(4)} & {Cat(18)} & {Cat(9)}\\
    \end{tabular}
    \label{table:ale}
\end{table*}

\subsubsection{Distillation Stopping Criteria}
When doing data-free policy distillation we measure student progress using a periodic evaluation. When the score on these evaluations stops improving for some amount of time or some number of new parameter versions we stop the experiment and save the student with the best recorded performance.

On ALE environments we stop experiments after 10,000 new parameter versions stop improving. On classical control environments we stop after 120 seconds without progress.

\subsubsection{Distillation Hyperparameters}

Student policies use neural network architectures with half the output size and filters of the teacher architectures given in Table~\ref{table:envhyperparams}. Students and generator use the Adam optimizer, and we use $\alpha=0.5$ and $\beta=5$ in the generator objective $\mathcal L_H$ \eqref{eq:generatorlossentropy}. Other GAS hyperparameters can be found in Table~\ref{table:rlgashparams}.

\begin{table}[H]
\caption{GAS Hyperparameters for Policy Distillation}
  \label{table:rlgashparams}
  \centering
  \begin{tabular}{lrrrrr}
  \toprule
  Environment & Student LR & Generator LR & Student Steps & Generator Steps & Generator Reset\\
  \midrule
  Humanoid & 0.001 & 0.001 & 5 & 2 & 10\\
  Standup & 0.001 & 0.001 & 5 & 2 & 10\\
MountainCar & 0.001 & 0.001 & 5 & 2 & 10\\
  Cartpole & 0.001 & 0.001 & 5 & 2 & 10\\
  ALE & 0.0003 & 0.1 & 7 & 5 & 256\\
  \end{tabular}
\end{table}

\subsection{Supervised Learning Teachers}
\label{appendix:supervisedteachers}
\subsubsection{MNIST and Fashion-MNIST}
Our MNIST and Fashion-MNIST teachers are trained using an Adam \cite{kingma2014adam} optimizer with learning rate 0.0007, and a Dropout \cite{srivastava2014dropout} probability of 0.2. We save the best performing model with a test accuracy of 99.35\% on MNIST and 91.17\% on Fashion-MNIST. We freeze the teachers and use them to compute generator and distillation losses. To be comparable to prior work we adopt a a LeNet-5 \cite{lecun2015lenet} architecture for the teacher, and the student has a LeNet-5-Half architecture (number of filters is halved). Our generator uses the DCGAN architecture from \cite{radford2015unsupervised}, which was also used in DAFL \cite{chen2019data}.

MNIST and Fashion-MNIST generators use the adaptive loss. sample a batch of 512 noise vectors $z \in \mathbb R^{100}$, $z_i \sim \mathcal U(-1, 1)$. These vectors are sliced into 64 sub-batches fed into each of the 8 concurrent generators. We reinitialize the generator once every 40,000 student updates on MNIST, and once every 120,000 student updates on Fashion-MNIST. After each 100 updates to the student, the generators are optimized for 20 steps. The generator and student are trained using the Adam \cite{kingma2014adam} optimizer, with a generator learning rate of 0.01 and a student learning rate of 0.001. We repeat our MNIST and Fashion-MNIST experiment with 3 seeds.

\subsubsection{CIFAR-10}

We use a ResNet-34 \cite{he2016deep} architecture as the teacher, and ResNet-18 \cite{he2016deep} as the student.
The student is trained using SGD with momentum 0.9, and a learning rate of 0.1 that is dropped by a factor of 10 every 96,000 gradient steps, for a total of 240,000 steps. The generator is trained using the Adam \cite{kingma2014adam} optimizer with a learning rate of 0.015. We perform 1 generator update for each 10 student updates. We set $\gamma = 0.4$ in the generator loss \eqref{eq:adaptiveloss}.

\end{document}